\documentclass[10pt,twocolumn,letterpaper]{article}

\usepackage[pagenumbers]{cvpr} 

\definecolor{cvprblue}{rgb}{0.21,0.49,0.74}
\usepackage[pagebackref,breaklinks,colorlinks,allcolors=cvprblue]{hyperref}
\usepackage{multirow}
\usepackage{colortbl}

\def\paperID{*****} 
\def\confName{ICCV}
\def\confYear{2025}

\title{‌CM3AE: A Unified RGB Frame and Event-Voxel/-Frame Pre-training Framework}

\author{
    Wentao Wu\textsuperscript{\rm 1,2,3}, 
    Xiao Wang\textsuperscript{\rm 1,2,4}\thanks{Corresponding author: Xiao Wang \& Chenglong Li}, 
    Chenglong Li\textsuperscript{\rm 1,2,3} 
    Bo Jiang\textsuperscript{\rm 1,2,4}, 
    Jin Tang\textsuperscript{\rm 1,2,4},  
    Bin Luo\textsuperscript{\rm 1,2,4}, 
    Qi Liu\textsuperscript{\rm 5} \\ 
${^1}$ \small{Information Materials and Intelligent Sensing Laboratory of Anhui Province, Anhui University, Hefei 230601, China} \\
${^2}$ \small{Anhui Provincial Key Laboratory of Multimodal Cognitive Computation, Anhui University, Hefei 230601, China} \\
${^3}$ \small{School of Artificial Intelligence, Anhui University, Hefei 230601, China} \\ 
${^4}$ \small{School of Computer Science and Technology, Anhui University, Hefei 230601, China} \\ 
${^5}$ \small{University of Science and Technology of China, Hefei 230601, China} \\ 
}

\begin{document}
\maketitle
\begin{abstract}
Event cameras have attracted increasing attention in recent years due to their advantages in high dynamic range, high temporal resolution, low power consumption, and low latency. 
Some researchers have begun exploring pre-training directly on event data. Nevertheless, these efforts often fail to establish strong connections with RGB frames, limiting their applicability in multi-modal fusion scenarios. To address these issues, we propose a novel CM3AE pre-training framework for the RGB-Event perception. This framework accepts multi-modalities/views of data as input, including RGB images, event images, and event voxels, providing robust support for both event-based and RGB-event fusion based downstream tasks. 
Specifically, we design a multi-modal fusion reconstruction module that reconstructs the original image from fused multi-modal features, explicitly enhancing the model's ability to aggregate cross-modal complementary information. Additionally, we employ a multi-modal contrastive learning strategy to align cross-modal feature representations in a shared latent space, which effectively enhances the model's capability for multi-modal understanding and capturing global dependencies. We construct a large-scale dataset containing 2,535,759 RGB-Event data pairs for the pre-training. 
Extensive experiments on five downstream tasks fully demonstrated the effectiveness of CM3AE. 
Source code and pre-trained models will be released on \url{https://github.com/Event-AHU/CM3AE}. 
\end{abstract}    
\section{Introduction}
\label{sec:intro}

\begin{figure*}[!htp]
\center
\includegraphics[width=\textwidth]{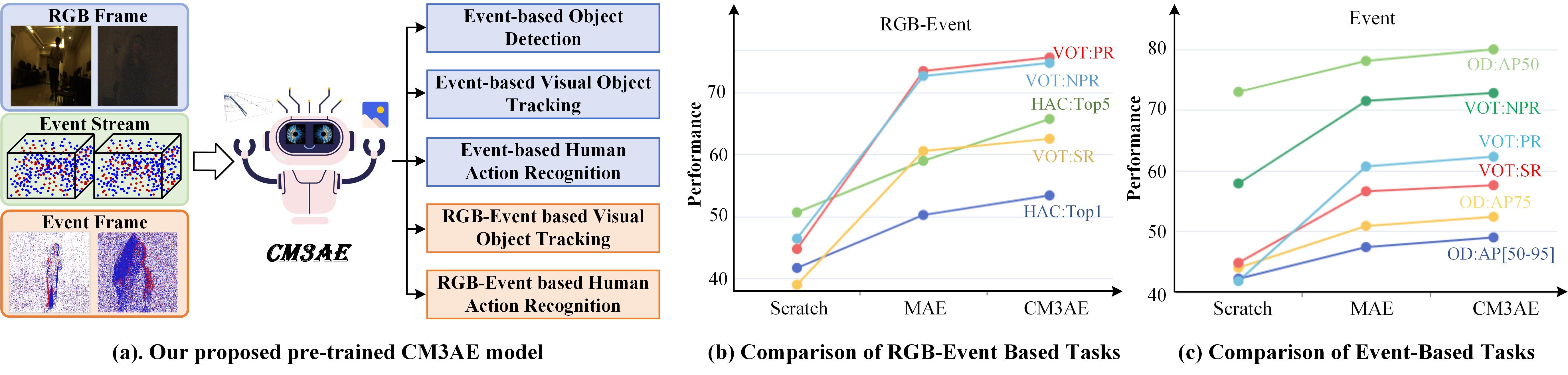}
\caption{Our proposed pre-trained CM3AE model takes RGB-Event Streams/-Frames as the input and supports unimodal and multimodal based downstream tasks, such as object detection and tracking, action recognition, etc. } 
\label{fig:firstimg}
\end{figure*}

Event cameras are neuromorphic sensors that offer advantages such as high temporal resolution, high dynamic range, low power consumption, and low latency. In recent years, it has been widely used in tasks such as object detection and tracking~\cite{mitrokhin2018event, wang2024event}, action recognition~\cite{wang2024hardvs}, scene understanding~\cite{kong2024openess}, and semantic segmentation~\cite{chen2024segment}. Early deep learning methods often conducted research on different tasks independently. Specifically, researchers typically constructed small task-specific datasets, trained models based on a generic pre-trained backbone network, and then evaluated them on a test subset. Although these approaches achieved performance improvements over traditional methods, many challenges remain unresolved due to the complexity of real-world events.

In recent years, large models have demonstrated remarkable performance across various tasks. They undergo task-agnostic pre-training on large-scale datasets, then fine-tuning on task-specific small-scale data, showcasing significant advantages in both accuracy and generalization. 
Specifically, large language models (LLMs), such as the GPT series~\cite{radford2018GPT1, radford2019GPT2, brown2020GPT3}, LLaMA series~\cite{touvron2023llama, touvron2023llama2, grattafiori2024llama}, and DeepSeek series~\cite{bi2024deepseek, dai2024deepseekmoe, liu2024deepseek, guo2025deepseek}, leverage knowledge learned from massive datasets to exhibit strong capabilities across diverse text-based tasks. In visual tasks, general pre-trained models such as the DINO series~\cite{zhang2022dino, oquab2024dinov2}, MAE~\cite{he2022mae}, and IBOT~\cite{zhouimage} have been widely applied to tasks like detection, recognition, and segmentation, achieving outstanding performance. At the same time, some large models designed for specific tasks, such as SAM~\cite{kirillov2023segment} and SEEM~\cite{zou2023segment}, are capable of performing arbitrary object segmentation in complex scenes. To obtain a more comprehensive and accurate model, researchers have attempted to use multimodal data for pre-training, such as RGB, natural language, audio, depth data, and thermal data. Foundation models like CLIP~\cite{radford2021learning}, MultiMAE~\cite{bachmann2022multimae}, and LLaVA~\cite{liu2023visual} have effectively improved the performance of multimodal tasks, and even in cases of modality missing, their performance remains strong compared to single-modal models.

Inspired by the success of the aforementioned large models, we begin exploring a multimodal pre-training framework based on event cameras, as shown in Fig.~\ref{fig:firstimg}. 
Previously, Yang et al.~\cite{yang2023event} augment event images into sample pairs and perform pre-training using contrastive learning. Huang et al.~\cite{huang2024data} propose a voxel-based pre-training framework that decomposes masked modeling process into local spatio-temporal and global semantic reconstruction. Klenk et al.~\cite{klenk2024masked} introduce a multi-stage framework that sequentially uses event histograms and raw images as reconstruction targets for pre-training. However, this approach makes it difficult to apply the model to Event-based multimodal tasks. In this study, we adopt the Masked Autoencoder (MAE) pre-training algorithm, which applies a high masking ratio to the input images and leverages an encoder-decoder structure to learn the reconstruction of the missing parts. However, simply extending MAE to a dual-branch structure effectively models individual modalities but overlooks cross-modal correlations, limiting its performance in multimodal tasks. 

Based on the aforementioned analysis, we believe that the framework can be optimized in the following aspects: 
\textbf{Firstly}, there exists complementary information between different modalities, and effectively integrating this complementary information is crucial for multimodal tasks. Therefore, during the pre-training phase, leveraging a large-scale paired dataset to train a robust fusion module can significantly enhance the model's performance on downstream multimodal tasks.
\textbf{Secondly}, the reconstruction process in masked image modeling primarily focuses on local relationships, which limits the model’s ability to learn global information. Meanwhile, contrastive learning is more effective in capturing global dependencies. Therefore, we introduce contrastive learning between different modality inputs to enhance the model’s capability in modeling global information and enhancing cross-modal understanding. 

In this paper, we propose a general RGB-Event multimodal pre-training framework building upon MAE. It introduces a fusion reconstruction module and a multimodal contrastive learning strategy to better explore the correlation between multimodal features, thereby enhancing the model's understanding of multimodal information. As shown in Fig.~\ref{fig:framework}, the proposed CM3AE consists of three components, i.e., the dual-branch masked autoencoder, the multimodal fusion reconstruction module, and multimodal contrastive learning module. 
Specifically, we first divide the input RGB and Event images into non-overlapping image patches, mask 75$\%$ of the patches, and then send the remaining patches into modality-specific Transformer encoders for feature representation learning. Afterward, the decoder is used to predict the pixel values of the masked image patches. In the fusion reconstruction module, we encode the Event voxel using the voxel encoder and combine it with the features from the RGB and Event encoders to feed into the fusion module. In the fusion module, we perform fusion using two combinations, i.e., RGB+Event and RGB+Event+Voxel, and use the fused features to reconstruct the original RGB image. At the same time, we treat different modalities of the same image as positive samples and other images as negative samples. We perform multimodal contrastive learning between RGB-Event and RGB-Voxel pairs to enhance the model's ability to extract global information. Extensive experiments on five downstream tasks demonstrate the effectiveness of our proposed CM3AE foundation model.

To sum up, we draw the main contributions of this paper as the following three aspects: 

$\bullet$ We propose a framework for RGB frame and event-voxel/-frame based pre-training, termed CM3AE. 
 It incorporates a multimodal fusion reconstruction module and a multimodal contrastive learning strategy to enhance the model's capability to understand multimodal information. 

$\bullet$ We construct a large-scale RGB-Event dataset for multimodal pre-training, termed REV2M, which contains 2,535,759 RGB-Event data pairs. 

$\bullet$ We conduct extensive experiments on five downstream tasks to validate the effectiveness of the proposed CM3AE framework, including RGB-Event based human action recognition and visual object tracking, as well as Event-based object detection, human action recognition, and visual object tracking. 

\section{Related Works}
\label{sec:related}

\subsection{Pre-trained Large Models}

Building large-scale datasets and training a foundational model through self-supervised/unsupervised methods is a popular research topic. Currently, the mainstream approaches for pre-trained models can be categorized into two types: contrastive learning based and reconstruction-based models.
Specifically, contrastive learning methods primarily construct representations by learning how to differentiate between input pairs and then encoding them. For unimodal pre-training, sample pairs are often created through data augmentation to facilitate contrastive learning. 
For instance, MoCo~\cite{he2020momentum} introduced the concept of a momentum encoder to address the limitation in the number of negative samples in contrastive learning.
In the context of multimodal pre-training, data from different modalities naturally form sample pairs, and contrastive learning is achieved by aligning these modalities. 
To be more precise, CLIP~\cite{radford2021learning} is pre-trained using image-text pairs and has demonstrated remarkable performance across various downstream tasks. 
Florence~\cite{yuan2021florence} aims to expand the scope of representations, ranging from coarse to fine and from static to dynamic, demonstrating even more exceptional performance.

Reconstruction-based pre-training methods achieve self-supervision by reconstructing the masked portions of the input within the model. 
Representative works, such as MAE~\cite{he2022mae} and BEiT~\cite{baobeit}, learn representations by predicting the original pixels of the masked regions in the input images. VideoMAE~\cite{tong2022videomae} and VideoMAE-v2~\cite{wang2023videomae} propose a dual-masking strategy, enabling more efficient pre-training for video objectives. 
Meanwhile, some researchers have applied the concept of masked modeling to pre-training for specific domains or targets. 
For example, SatMAE~\cite{cong2022satmae}, Scale-MAE~\cite{reed2023scale}, and SatMAE++~\cite{noman2024rethinking} focus on the remote sensing domain, HDXrayMAE\cite{wang2024HDXrayMAE} targets the medical field, HumanBench~\cite{tang2023humanbench}, UniHCP~\cite{ci2023unihcp}, and PLIP~\cite{zuo2023plip} are designed for human-related tasks, and VehicleMAE~\cite{wang2024structural} addresses vehicle-related objectives.
Unlike existing pre-training research focused on general scenes, we propose a large pre-training model specifically tailored for visible light and event cameras. Our model is capable of enabling a variety of downstream tasks and applications related to event cameras.

\subsection{Multimodal Learning}

Multimodal learning aims to leverage data from different sources to enhance a model's perception and understanding capabilities. However, inherent modality differences exist among multimodal data, making it difficult to fuse complementary information across modalities. Most current research primarily focuses on eliminating these modality differences to better integrate multimodal complementary information, thereby improving the accuracy of corresponding tasks.
Li et al.~\cite{li2025mulfs} propose using a shared shallow feature encoder and learnable modality dictionaries, which effectively mitigates the impact of modality differences on cross-modal feature alignment and fusion.
Li et al.~\cite{licoupled}, in order to address the challenge of capturing cross-modal interaction dynamics under complex intra-modal and cross-modal associations, proposed a coupled state-space model that maintains the independence of intra-modal states while coupling cross-modal state chains.
Chen et al.~\cite{chen2024weakly} proposed an offset-guided adaptive feature alignment method (OAFA) to address the spatial misalignment issue between different modality images, enabling the adaptive adjustment of the relative positions of multimodal features.
Some researchers attempt to perform pretraining on large-scale multimodal data to better integrate multimodal information. MultiMAE~\cite{bachmann2022multimae} proposes a pretraining framework based on multimodal multi-task masked autoencoders, achieving promising performance in both unimodal and multimodal downstream tasks. Gang et al.~\cite{geng2022multimodal} propose a multimodal masked autoencoder that learns a unified encoder for vision and language data through masked token prediction, demonstrating strong transfer capabilities on vision-language downstream tasks. Inspired by these works, we fully leverage large-scale multimodal paired data during pretraining to explore inter-modal correlations and enhance the model's understanding of multimodal data.

\begin{figure*}[!htp]
\center
\includegraphics[width=\textwidth]{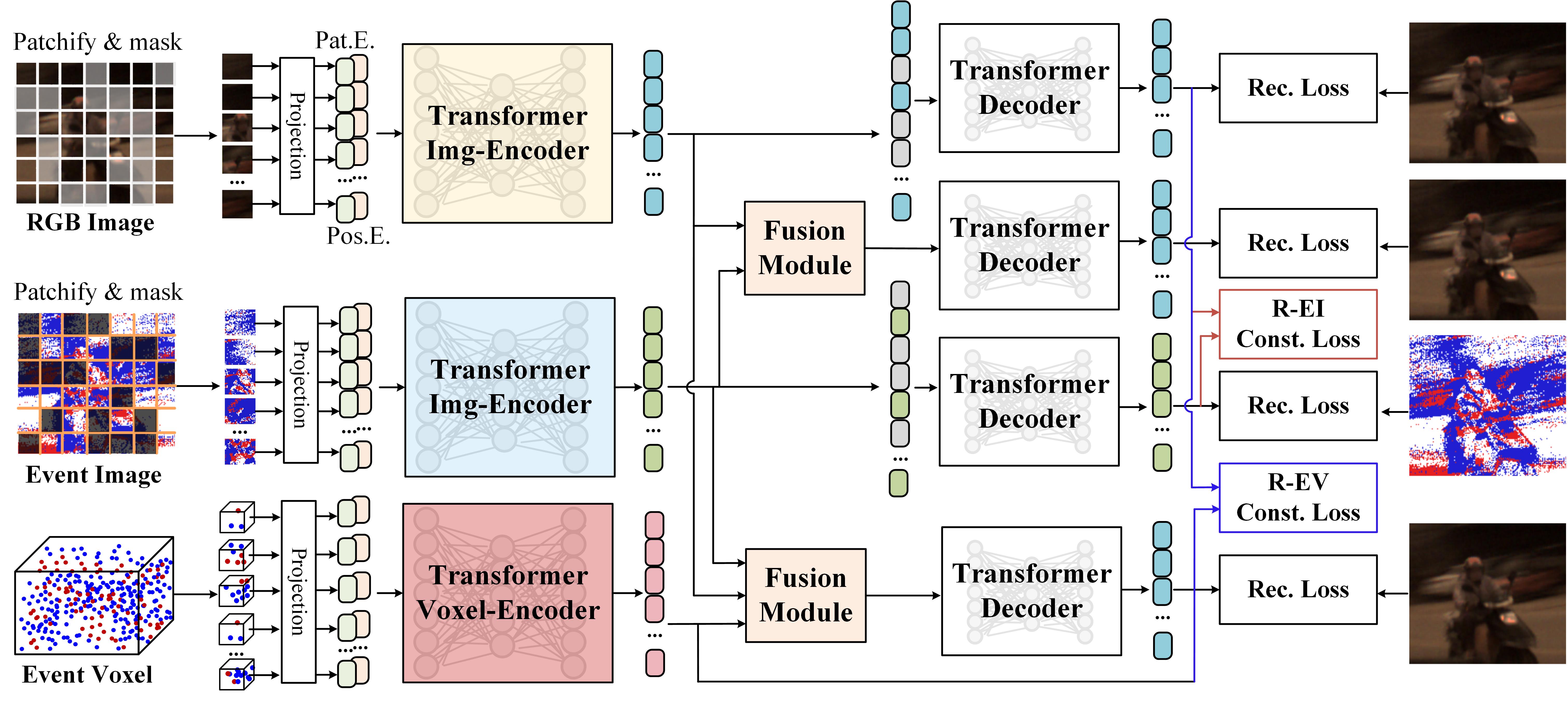}
\caption{An overview of our proposed pre-training framework for RGB-Event stream perceptron, termed CM3AE. Specifically, the framework supports multi-modalities/views of data as input, incorporating a multi-modal fusion generation module and a multi-modal contrastive learning strategy. These designs effectively enhance the model's ability to aggregate cross-modal information and improve multi-modal understanding, significantly boosting performance across various event-based single-modal tasks and RGB-Event fused multi-modal downstream tasks.} 
\label{fig:framework}
\end{figure*}

\subsection{Event Stream based Downstream Tasks}
In recent years, many researchers have dedicated their efforts to studying vision problems based on event streams. Among these, the most extensively researched visual tasks include object detection and action recognition.  
In the object detection task, 
SpikingYOLO~\cite{kim2020spiking} combines YOLO with Spiking Neural Networks (SNN) and introduces the first object detection model that utilizes deep SNN. SpikeYOLO~\cite{luo2024integer} proposes the I-LIF neuron, which reduces quantization errors in SNN by training with integer values and performing inference as a binary spike sequence. 
SFOD~\cite{fan2024sfod} achieves multi-scale feature map fusion for the first time in SNN applied to event cameras, enhancing the model's detection performance. 
Wang et.al~\cite{wang2024object} proposed MOE-HCO blocks integrate multiple expert models to mimic the heat conduction process in event streams, effectively balancing accuracy and computational efficiency. Additionally, by transforming event streams into image form, it can also be directly applied to RGB image-based object detection models, such as DetectoRS~\cite{qiao2021detectors}, DETR~\cite{carion2020end}, and ViTDet~\cite{li2022exploring}.

In the human action recognition task, 
Wu et al.~\cite{wu2020multipath} generate event frames by accumulating asynchronous time and improve recognition accuracy by integrating the complementary information from event frames and human pose. 
Wang et al.~\cite{wang2024hardvs} propose a spatiotemporal feature learning and fusion behavior recognition framework, ESTF, which enhances recognition accuracy by fusing RGB and event dual-view features. 
ExACT~\cite{zhou2024exact} introduced an adaptive fine-grained event representation and uncertainty estimation module based on conceptual reasoning, performing event-based action recognition from the perspective of cross-modal conceptualization. 
HyperMV~\cite{gao2024hypergraph}, in order to fully utilize multi-view event data, constructs a multi-view hypergraph neural network to capture the relationships between event features across different views of the target.
These works generally use ImageNet pre-trained weights to initialize their network weights. In this paper, we propose a novel potential approach for pre-training large multimodal models, providing a solid foundation for subsequent event stream perception and multi-modal fusion tasks.

\section{Methodology}
\label{sec:meth}

\subsection{Overview} 

As shown in Fig.~\ref{fig:framework}, the proposed CM3AE follows a dual-branch masked auto-encoder framework. Specifically, we first divide paired RGB and Event images into non-overlapping image patches, and then apply high-proportion random token masking to them. We ensure that half of the unmasked tokens from the RGB and Event modalities correspond to the same positions. The unmasked tokens are then fed into the network. In this paper, we use a Transformer encoder-decoder structure to reconstruct the masked tokens. More importantly, to better model the relationship between the two modalities, we design a multimodal fusion reconstruction module and a multimodal contrastive learning strategy. For the fusion reconstruction module, we fuse the features output by the encoder in two combination modes: RGB+Event and RGB+Event+Voxel. The fusion is constrained by reconstructing the original RGB image. Since the multimodal data in the pre-training dataset are all collected in pairs, we introduce multi-model contrastive losses for RGB-Event and RGB-Voxel to construct a cross-modal shared semantic space, mitigate modal alignment bias, and simultaneously enhance the model's ability to extract global information. Experiments on multiple downstream tasks show that the multimodal fusion reconstruction module and multimodal contrastive learning significantly improve the pre-training performance.

\subsection{Network Architecture} 

Our proposed CM3AE contains three main parts, i.e., dual-branch masked auto-encoder(DMA), multimodal fusion reconstruction module(MFRM), and multimodel contrastive learning(MCL).

\noindent
\textbf{Dual-branch Masked Auto-Encoder.}
Given an input pair of RGB-Event images $I_{RGB}$ and $I_{Event} \in \mathbb{R}^{224 \times 224 \times 3}$, we divide them into 196 non-overlapping image patches $P_i \in \mathbb{R}^{16 \times 16 \times 3}, i \in \{1, 2, ... , 196\}$ using the same approach. Before feeding them into the encoder, we randomly mask 75$\%$ of the patches. Unlike other masked image modeling methods, we ensure that half of the unmasked patches in the RGB and Event modalities are identical. This design prevents the multimodal fusion reconstruction module from leaking information about the masked patches. The unmasked image patches pass through a convolutional layer with kernels $16 \times 16$ and are projected into the token embeddings $E_j \in \mathbb{R}^{1 \times 768}, j \in \{1, 2, ..., 49\}$. Similar to other works, we add the CLS-token before token embedding, obtaining input token embeddings $E_{RGB} \in \mathbb{R}^{50 \times 768}$ and $E_{Event} \in \mathbb{R}^{50 \times 768}$. Meanwhile, we randomly initialize positional encoding $Z^1_{RGB}$ and $ Z^1_{Event} \in \mathbb{R}^{50 \times 768}$ to represent the spatial position of tokens and add it to the token embeddings, resulting in the final input embeddings $\tilde{E}_{RGB}$ and $\tilde{E}_{Event}$.

Then, we feed the final input embeddings $\tilde{E}_{RGB}$ and $\tilde{E}_{Event}$ of the two modalities into the corresponding ViT-B/16 encoders. Each encoder consists of 12 Transformer modules, with each module comprising layer normalization, multi-head self-attention (MSA), and a multi-layer perceptron (MLP). The outputs of the Transformer encoders for each modality are projected through a 512-dimensional linear projection layer, mapping them to the visible token embeddings $\bar{E}_{RGB}$ and $\bar{E}_{Event} \in \mathbb{R}^{50 \times 512}$. Before inputting to the decoder, we construct a set of mask tokens $E^{m}_{RGB}$ and $E^{m}_{Event} \in \mathbb{R}^{147 \times 512}$ for each modality, which are shared and learnable vectors. Additionally, a positional encoding $Z^2_{RGB}$ and $ Z^2_{Event} \in \mathbb{R}^{197 \times 512}$ is introduced and combined with the mask tokens. After concatenating the mask tokens and visible tokens for each modality, they serve as the inputs $\hat{E}_{RGB} \in \mathbb{R}^{197 \times 512}$ and $\hat{E}_{Event} \in \mathbb{R}^{197 \times 512}$ to the decoder. The decoder network consists of 8 Transformer modules, which are used only during the pre-training phase for image reconstruction.

The mean square error (MSE) over pixel space from the masked token in the original image and the reconstructed token is adopted as the reconstruction loss to optimize the MAE module, which can be written as:
\begin{equation}
L_{m} = \frac {1} {N_{m}} (\sum_{t_{1}\in P_{rm}} || V^{r}_{t_{1}} - \bar{V}^{r}_{t_{1}} ||_{2} + \sum_{t_{2}\in P_{em}} || V^{e}_{t_{2}} - \bar{V}^{e}_{t_{2}} ||_{2}),
\end{equation}
where $V^{r}$ and $V^{e}$ represent the pixel values of the input RGB and Event images, respectively, $\bar{V}^{r}$ and $\bar{V}^{e}$ represent the predicted pixel values for RGB and Event images. $N_{m}$ is the number of masked pixels, $P_{rm}$ and $P_{em}$ are the indices of the RGB and Event masked pixels, $||*||_2$ refers to the $L_{2}$ loss.  

\noindent 
\textbf{Multimodal Fusion Reconstruction Module.}
The aforementioned DMA module has already shown strong performance in RGB-Event-based visual tasks, but it still lacks modeling the relationships between the two modalities, which is crucial for multimodal tasks. Therefore, we design a Multimodal Fusion Reconstruction Module(MFRM), which integrates multimodal features and reconstructs the original images. This approach helps the model better capture the correlations between modalities. At the same time, the fusion module within this module can be directly transferred to downstream tasks, effectively improving their performance.

In this module, we first select tokens from the Event embeddings $\bar{E}_{Event}$ output by the encoder that correspond to the same positions in the RGB embeddings $\bar{E}_{RGB}$. Then, we concatenate them with the RGB token embeddings $\bar{E}_{RGB}$ to obtain the input $\tilde{E}_{re} \in \mathbb{R}^{75 \times 512}$ for the fusion module. The fusion module is a standard Transformer block, consisting of three components: layer normalization, multi-head self-attention, and a multi-layer perceptron. We remove the Event tokens from the fusion module output $\bar{E}_{re}$, and then concatenate them with the masked tokens $E^{m}_{re} \in \mathbb{R}^{147 \times 512}$ in the same way as described above. The resulting embedding $E_{f}$ are fed into the decoder for reconstructing the original RGB image.

At the same time, in another branch of the MFRM, we also introduce event voxel data. Since voxel data is three-dimensional, the same patch division method used for 2D images cannot be directly applied. In this paper, each voxel contains 14 points, and for each image, we select a fixed number of voxels. If the number of voxels in the image is lower than the fixed value, we sample from the existing voxels to fill it. If the number exceeds the fixed value, we randomly sample to obtain the fixed number. Then, we divide the voxels into multiple groups, with the number of groups matching the number of image patches. The voxels within each group are concatenated into a single token. After this, we pass the resulting voxel tokens through a 768-dimensional linear layer to obtain the voxel token embedding $E_{vox} \in \mathbb{R}^{196 \times 768}$. Then, we add a CLS-token before the voxel token embedding, and after adding positional encoding, we input it into the encoder, which consists of 12 stacked residual attention blocks. We pass the output of the voxel encoder $E_{vox}$, through a 512-dimensional linear layer to reduce its dimensionality, and then concatenate it with the aforementioned $\tilde{E}_{re}$ , obtaining the input $\tilde{E}_{rev} \in \mathbb{R}^{75 \times 512}$ for the fusion module. For the fused features, we retain only the RGB token embeddings, and after concatenating the mask token, feed them into the decoder to reconstruct the original RGB image.

In the MFRM, we still use MSE as the reconstruction loss. The overall loss of the module can be expressed as:
\begin{equation}
L_{f} = \frac {1} {N_{m}} (\sum_{t\in P_{rm}} || V^{r}_{t} - \bar{V}^{re}_{t} ||_{2} + \sum_{t\in P_{rm}} || V^{r}_{t} - \bar{V}^{rev}_{t} ||_{2}),
\end{equation}
where $\bar{V}^{re}$ represents the predicted pixel values of the RGB image from the fused RGB and Event features, and $\bar{V}^{rev}$ represents the predicted pixel values of the RGB image from the fused RGB, Event, and voxel features.

\begin{figure*}[!htp]
\center
\includegraphics[width=\textwidth]{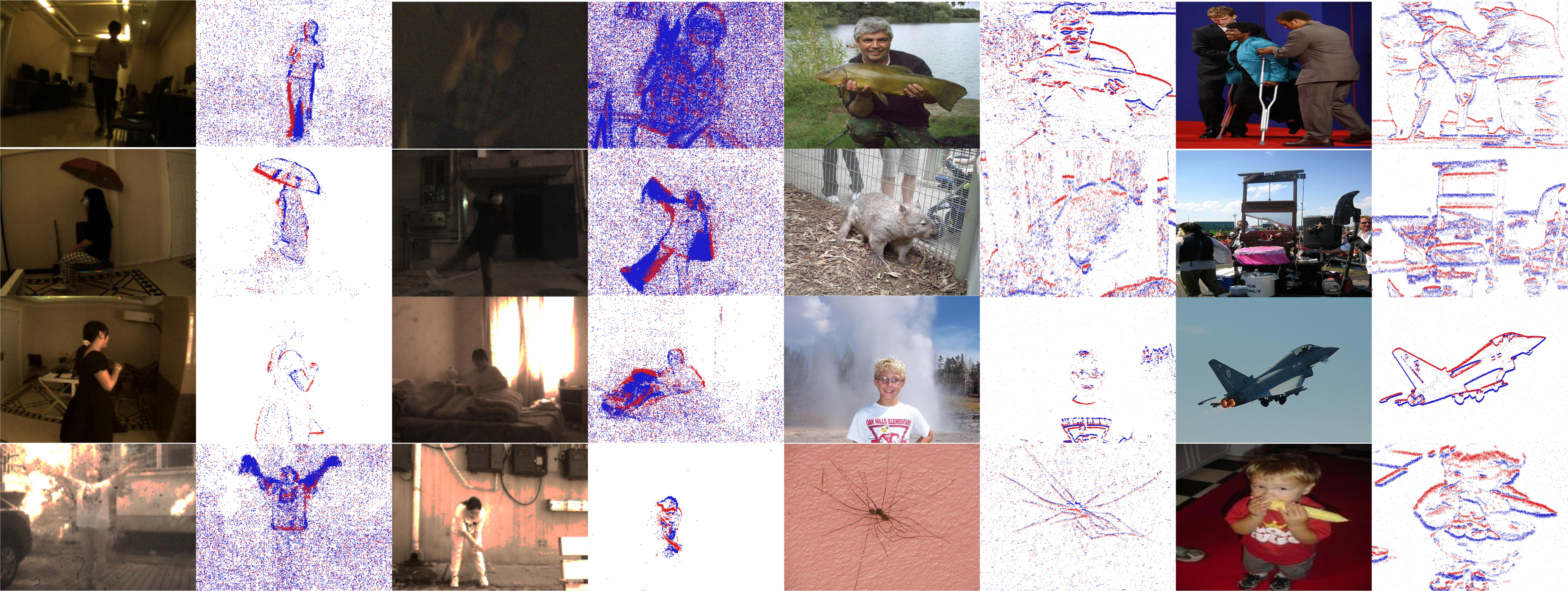}
\caption{Representative samples in our pre-training dataset.} 
\label{dataSamples}
\end{figure*}

\noindent 
\textbf{Multimodal Contrastive Learning.}
We first reconstruct the original images of each modality in an autoregressive manner to train the encoder to better extract the representations of each modality. Based on this, we propose a MFRM that uses the fused features to generate the original images, effectively mining the latent correlations between different modalities. Although this approach enhances the model's performance, the reconstruction process mainly focuses on modeling local relationships, leading to certain limitations when handling global dependencies. Therefore, we introduce the concept of multimodal contrastive learning, encouraging the model to better understand and integrate multimodal information, thereby improving its ability to capture global dependencies.

In this work, we use a pre-training dataset containing paired RGB, Event, and Voxel data. Therefore, we perform contrastive learning between RGB and Event as well as between RGB and Voxel. Taking the contrastive learning process between RGB and Event as an example, we first perform $L_2$ normalization on the reconstructed features $F_{RGB}$ and $F_{Event}$ output by the RGB and Event decoders. The normalization process is as follows:
\begin{equation}
\tilde{F}_{RGB} = \frac {F_{RGB}} {||F_{RGB}||_{2}} , 
\quad
\tilde{F}_{Event} = \frac {F_{Event}} {||F_{Event}||_{2}}.
\end{equation}
Then, we compute the logits for each modality using the normalized features. The calculation formula is:
\begin{equation}
lg_{re} = ls \cdot \tilde{F}_{RGB} \cdot (\tilde{F}_{Event})^T,  
\quad
 lg_{er} = ls \cdot \tilde{F}_{Event} \cdot (\tilde{F}_{RGB})^T,
\end{equation}
where $ls$ is the scaling factor. When computing the loss, we consider different modalities of the same image as positive samples, while all other images are treated as negative samples. Therefore, the loss for RGB and Event contrastive learning can be expressed as:
\begin{equation}
L_{re} = - \frac{1}{N}  \sum_{i=1}^{N} \log \left( \frac{\exp(lg^{i}_{re})}{\sum_{j=1}^{N} \exp(lg^{j}_{re})} \right),
\end{equation}
\begin{equation}
L_{er} = - \frac{1}{N}  \sum_{i=1}^{N} \log \left( \frac{\exp(lg^{i}_{er})}{\sum_{j=1}^{N} \exp(lg^{j}_{er})} \right),
\end{equation}
where $N$ represents the number of images of a single modality in a batch. The total loss $L_{cl}$ for MCL can be expressed as:
\begin{equation}
L_{cl} = L_{re} + L_{er} + L_{rv} + L_{vr},
\end{equation}
where $L_{rv}$ and $L_{vr}$ are the losses during the contrastive learning between RGB and Voxel, calculated in the same way as described above. Therefore, the total loss for the model training process is:
\begin{equation}
L = L_{m} + L_{f} + L_{cl}.
\end{equation}

\subsection{Downstream Tasks} 

In this work, five downstream tasks are adopted to validate the effectiveness and generalization of our proposed CM3AE large model, including Event-/RGB-Event based human action recognition~\cite{bertasius2021space}, Event-based object detection~\cite{li2022exploring}, and Event-/RGB-Event based visual object tracking~\cite{zheng2024odtrack,tang2022revisiting}. A brief introduction to these tasks can be found in the supplementary materials.

\begin{table*}
\centering
\caption{Experimental results of ours and other pre-trained models on action recognition, object detection and visual object tracking.}
\label{tab:result_all} 
  \resizebox{\textwidth}{!}{
  \begin{tabular}{c|c|cc|cc|ccc|ccc|ccc}
    \hline  \toprule
    \multirow{3}{*}{\raggedright \textbf{Method}} & \multirow{3}{*}{\raggedright \textbf{Dataset}} & \multicolumn{4}{c|}{\textbf{Action Recognition}} & \multicolumn{3}{c|}{\textbf{Object Detection}}  & \multicolumn{6}{c}{\textbf{Visual Object Tracking}}  \\
    \cline{3-6} \cline{7-9} \cline{10-15}
    &  & \multicolumn{2}{c|}{\textbf{RGB+Event}} &\multicolumn{2}{c|}{\textbf{Event}} & \multicolumn{3}{c|}{\textbf{Event}} & \multicolumn{3}{c|}{\textbf{RGB+Event}} &\multicolumn{3}{c}{\textbf{Event}} \\
    \cline{3-6} \cline{7-9} \cline{10-15}
     &  & Top1 & Top5 & Top1 & Top5 & $AP_{[0.5:0.95]}$  & $AP_{0.5}$ &  $AP_{0.75}$  & SR & PR & NPR & SR & PR & NPR \\
    \hline 
    Scratch & -   &41.63 &50.66   &20.23 &37.74    &42.2 &73.0 &44.0    &38.9 &44.7 &46.4   &44.8 &41.8 &57.9 \\
    
    BEiT~\cite{baobeit} & ImageNet-1K    &43.45 &51.75   &26.42 &43.43     &45.8 &77.6 &48.7   &41.8 &48.1 &51.1   &48.0 &47.1 &61.2 \\
    
    DINO~\cite{zhang2022dino} & ImageNet-1K    &48.30 &54.38   &31.99 &47.08    &46.7 &78.4 &48.2    &54.1 &65.3 &66.4   &55.7 &59.5 &70.7\\
    
    Mocov3~\cite{chen2021empirical} & ImageNet-1K    &44.09 &53.24   &25.77 &44.21    &47.2 &77.3 &49.5 
    &53.9 &61.1 &66.2   &54.2 &56.7 &69.3 \\
    
    IBOT~\cite{zhouimage} & ImageNet-1K    &45.12 &51.36   &22.99 &39.99    &46.2 &76.3 &49.1    &58.3 &71.3 &70.3   &55.7 &59.7 &70.6  \\
    
    BEiTv2~\cite{peng2022beit} & ImageNet-1K    &45.01 &51.76   &26.61 &43.52    &46.6 &78.4 &48.4    &43.1 &51.6 &53.1   &51.1 &52.0 &65.0  \\

    EVA~\cite{wang2023image} & ImageNet-1K    &48.14 &55.06   &36.44 &50.96    &47.5 &78.7 &50.7    &58.1 &70.7 & 70.5  &56.0 &60.3 &71.1\\

    PIXMIM~\cite{liupixmim} & ImageNet-1K    &47.17 &53.44   &35.07 &49.67       &47.3 &77.5 &50.7  &59.4 &72.8 &71.2   &55.6 &59.7 &70.5  \\

    MAE~\cite{he2022mae} & ImageNet-1K    &50.23 &59.65   &49.21 &59.01    &47.4 &78.1 &50.9    &60.6 &73.6 &72.8   &56.6 &60.7 &71.4  \\
    
    \hline 
    MAE~\cite{he2022mae} & REV2M    &50.98 &62.10   &49.32 &60.41   &47.5 &78.0 &50.3    &61.0 &73.8 &73.1   &56.8 &60.5 &71.5  \\
    \rowcolor[rgb]{0.95,0.95,0.95} Our & REV2M    &\textbf{53.40} &\textbf{65.82}   &\textbf{52.45} &\textbf{63.76}   &\textbf{49.0} &\textbf{80.0} &\textbf{52.4}    &\textbf{62.6} &\textbf{75.8} &\textbf{74.9}   &\textbf{57.6} &\textbf{62.3} &\textbf{72.8}  \\
    \bottomrule
  \end{tabular}  }
\end{table*}

\noindent $\bullet$ \textbf{Event-/RGB-Event based Human Action Recognition.~} 
This task aims to recognize human actions in input videos, such as drinking water, riding a bicycle, and skipping rope, using a deep learning network. In this paper, we validate our proposed CM3AE pre-training model under two settings: Event-based and RGB-Event based. For the Event-based approach, we directly employ the Timesformer~\cite{bertasius2021space} model. However, unlike single-modal recognition tasks, the RGB-Event based method seeks to fully integrate the complementary information from RGB and Event modalities for target action recognition. Therefore, we perform a simple extension of the Timesformer~\cite{bertasius2021space} to adapt it to multimodal data input.

\noindent $\bullet$ \textbf{Event-based Object Detection.~} 
This task aims to leverage the characteristics of the event stream data to address challenges such as low light conditions, blur and rapid movement in object detection tasks, improving detection performance. In this work, we validate the proposed CM3AE pre-training model based on the VitDet~\cite{li2022exploring} model.

\noindent $\bullet$ \textbf{Event-/RGB-Event based Visual Object Tracking.~}
The objective of this task is to continuously and accurately identify and locate specific targets of interest within a given video sequence. The targets to be tracked are selected in the first frame of the input video sequence. In this work, we validate the proposed CM3AE pre-training model on both the unimodal tracking algorithm ODTrack~\cite{zheng2024odtrack} based on Event data and the multimodal tracking algorithm CEUTrack~\cite{tang2022revisiting} based on RGB-Event data.
\section{Experiments}
\label{sec:exper}

\subsection{Datasets and Evaluation Metric} 
To train our proposed CM3AE model, we integrate existing public datasets to construct a large-scale pre-training dataset containing 2M RGB-Event pairs. In Fig.~\ref{dataSamples}, we show some representative samples from the dataset. Then, we evaluate the effectiveness and generalization ability on four datasets corresponding to five downstream tasks. A brief introduction to these datasets is given below. 

\noindent 
$\bullet$ \textbf{Pre-training Dataset.} 
In this paper, we construct a large-scale RGB-Event multimodal dataset, REV2M, for pre-training. This dataset contains 2,535,759 RGB-Event pairs, sourced from five publicly available datasets: HARDVS~\cite{wang2024hardvs}, N-ImageNet~\cite{kim2021n}, COESOT~\cite{tang2022revisiting}, Visevent~\cite{wang2023visevent} and DSEC-MOD~\cite{zhou2024event}. These datasets consist of 827,694, 1,281,166, 231,277, 185,127, and 10,495 data pairs, respectively.

\noindent 
$\bullet$ \textbf{Downstream Datasets. } 
We validated the effectiveness and generalization ability of our pre-trained model on four datasets across five different downstream tasks, including the HARDVS~\cite{wang2024hardvs}, EvDET200K~\cite{wang2024object}, COESOT~\cite{tang2022revisiting}, and the EventVOT~\cite{wang2024event} dataset.

\noindent 
\textbf{HARDVS} 
is a large-scale human activity recognition dataset that contains over 100,000 video clips recorded using the DAVIS346 camera, with each clip lasting approximately 5-10 seconds. It includes 300 categories of human activity, such as drinking water, cycling, sitting down, and washing hands. 

\noindent 
\textbf{EvDET200K}
is an event-based object detection dataset captured using the high-resolution Prophesee EVK4-HD event camera. It covers 10 distinct categories, contains 200,000 bounding boxes, and includes 10,054 samples, each lasting 2 to 5 seconds.

\noindent 
\textbf{COESOT}
is a large-scale multimodal object tracking dataset based on RGB-Event, where videos are captured using a DVS346 event camera equipped with a zoom lens in diverse indoor and outdoor scenes. It includes 90 categories and 1,354 video sequences, with the training set and test set containing 827 and 527 videos, respectively.

\noindent 
\textbf{EventVOT}
is a high-resolution object tracking dataset based on Event, where all videos are captured using an EVK4-HD event camera, with an output event stream resolution of $1280 \times 720$. The dataset contains 1,141 video sequences, covering 19 categories including vehicles, drones, pedestrians, and more.

\noindent 
$\bullet$ \textbf{Evaluation Metric.}
In this work, multiple evaluation metrics are used for different downstream tasks, including Accuracy Top1, Accuracy Top5, $AP_{[0.5:0.95]}$, $AP_{0.5}$, $AP_{0.75}$, $SR$, $PR$, and $NPR$.

\begin{figure*}
\center
\includegraphics[width=\textwidth]{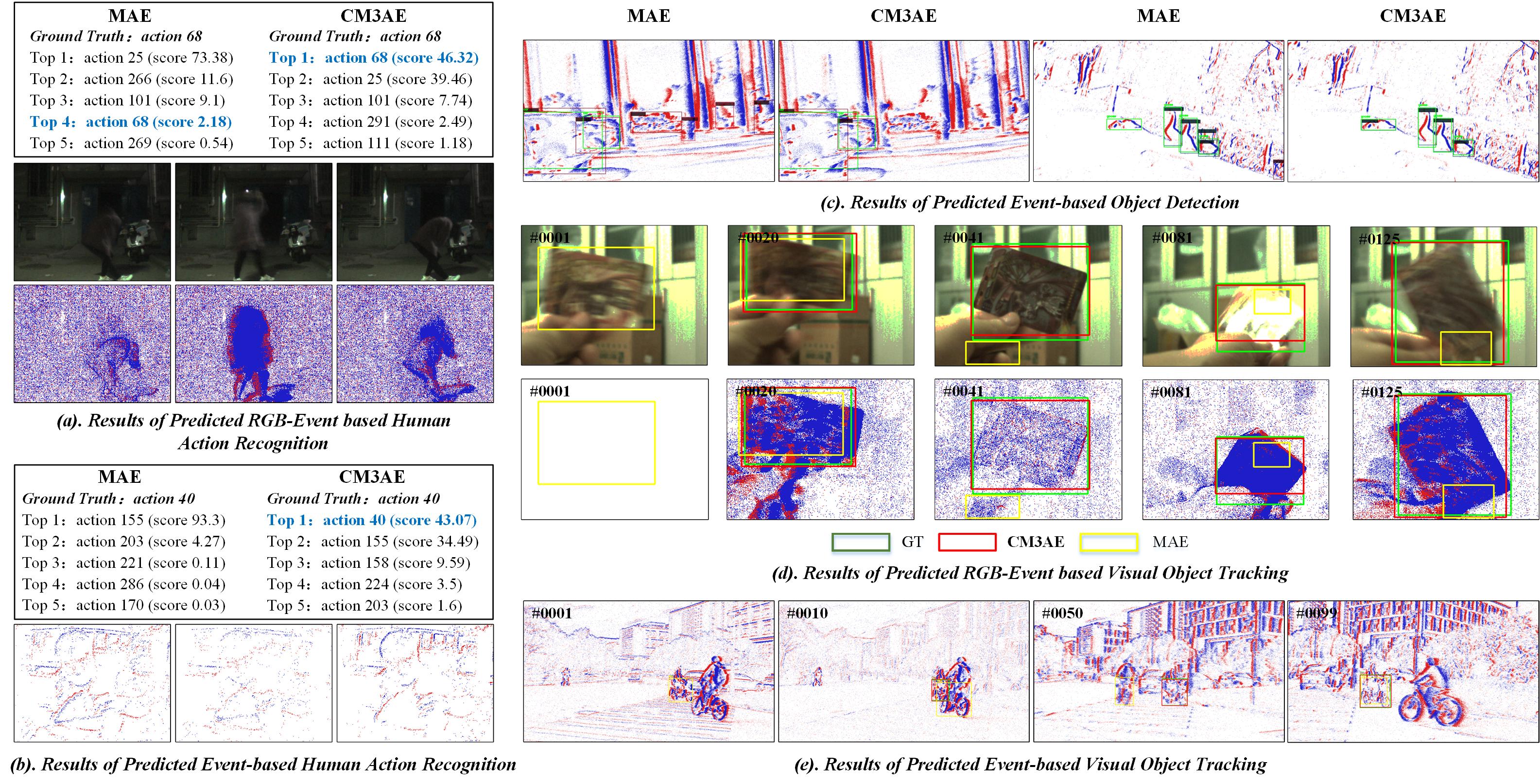}
\caption{Visualization of the experimental results of each downstream task.}    
\label{fig:downstream_vis}
\end{figure*}

\subsection{Implementation Details}  

In our pre-training phase, the learning rate is set to 0.0002, and the weight decay is 0.04. The AdamW~\cite{adamW} is selected as the optimizer to train our model. The batch size is 512 and the model is trained for a total of 200 epochs on our dataset.
All the experiments are implemented using Python based on the PyTorch~\cite{paszke2019pytorch}. A server with four A800 GPUs is used for the pre-training. About 134 hours are needed for our pre-training phase.

During the fine-tuning phase of downstream tasks, we configured distinct training parameters for different tasks. Specifically, for both Event-based and RGB-Event-based human action recognition tasks, we employed identical parameters: the models were trained for 15 epochs with a learning rate of 0.008 and a batch size of 8. For the Event-based object detection task, the input image size was set to 1024×1024, with training conducted for 15 epochs using a learning rate of 0.0001 and a batch size of 8. For the Event-based object tracking task, the learning rate was set to 0.0001 with a batch size of 16, and the model was trained for 50 epochs. For the RGB-Event-based object tracking task, the pre-trained model was fine-tuned with a learning rate of 0.0001, a batch size of 16, and trained for 100 epochs. The parameter settings for downstream tasks were largely consistent with their corresponding algorithms. Additionally, the training of Event-based object detection and tracking models was performed on two NVIDIA RTX 4090 GPUs, while the RGB-Event-based object tracking model was trained and tested on a single NVIDIA RTX 4090 GPU. All human action recognition tasks were trained and tested on an NVIDIA RTX 3090 GPU.

\subsection{Comparison on Public Benchmarks} 

In this experiment, we use five downstream tasks to validate our CM3AE pre-trained large model. We compare it with our baseline model and other state-of-the-art models, as shown in Table~\ref{tab:result_all}. Specifically, we compare the model without pre-training (i.e., learning from scratch), BEiT~\cite{baobeit}, DINO~\cite{zhang2022dino}, MoCov3~\cite{chen2021empirical}, IBOT~\cite{zhouimage}, BEiTv2~\cite{peng2022beit}, EVA~\cite{wang2023image}, PIXMIM~\cite{liupixmim}, and MAE~\cite{he2022mae} pre-trained on the ImageNet dataset, as well as the MAE model trained on our integrated pre-training dataset.

For the action recognition task, we conduct experiments on the HARDVS dataset. We selected Timesformer~\cite{bertasius2021space} as the baseline method and integrated the pre-trained model to validate the effectiveness of our approach. When the pre-trained model is not used, the baseline method achieves Top-1 and Top-5 accuracy of $20.23\%$ and $37.74\%$, respectively, in the Event single-modal setting, while in the RGB-Event multimodal setting, the Top-1 and Top-5 accuracy improve to $41.63\%$ and $50.66\%$, respectively. When using the parameters pre-trained by MAE on the ImageNet dataset to initialize the ViT-base backbone, the recognition accuracy in the Event single-modal setting improves to $49.21\%$ and $59.01\%$ for Top-1 and Top-5, respectively, while the RGB-Event multimodal setting achieves $50.23\%$ and $59.65\%$, respectively. This result indicates that the task-agnostic representations learned through the self-supervised process significantly contribute to downstream tasks. 

\begin{table}
  \centering
  \caption{Ablation study the multimodal fusion reconstruction module and multimodal contrastive learning.}
  \label{tab:ablation_model}
  \resizebox{3.3in}{!}{
  \begin{tabular}{ccc|c|cc|cc}
    \hline  \toprule
    \multirow{3}{*}{\raggedright \textbf{DMA}} &  \multirow{3}{*}{\raggedright \textbf{MFRM}} &\multirow{3}{*}{\raggedright \textbf{MCL}}  
    &\multirow{3}{*}{\textbf{Dataset}}
    &\multicolumn{4}{c}{\textbf{Action Recognition}} \\ 
    \cline{5-8}
     & & & &\multicolumn{2}{c|}{\textbf{RGB+Event}} &\multicolumn{2}{c}{\textbf{Event}} \\
    \cline{5-8}
     & & & & Top1 & Top5 & Top1 & Top5 \\
    \hline
    \checkmark & &  &\multirow{4}{*}{\raggedright \textbf{HARDVS}}  &51.06 &61.50   &49.47 &60.27  \\
    \checkmark & \checkmark &  &  &51.96 &63.46   &52.20 &62.03   \\
    \checkmark & & \checkmark  &  &51.62 &62.43   &51.49 &61.16   \\
    \checkmark & \checkmark & \checkmark &  &\textbf{53.24} &\textbf{65.13}   &\textbf{53.18} &\textbf{62.88}  \\

    \bottomrule
  \end{tabular} }
\end{table}

On this basis, we extend MAE to dual-modal input and conduct pre-training on the REV2M dataset. Compared to the original MAE, our approach achieves superior performance with only $25\%$ of the training time. This demonstrates that the performance of general pre-trained models is limited in event-camera-based downstream tasks. Notably, when using the pre-training method proposed in this paper, the best performance is achieved across all metrics, with Event single-modal recognition accuracy reaching $53.18\%$ and $62.88\%$, and RGB-Event multimodal recognition accuracy improving to $53.24\%$ and $65.13\%$. In contrast, existing pre-trained models such as BEiTv2, EVA, and PIXMIM all exhibited lower performance than our model. Similarly, the same trend was observed in the object detection task. These results and comparisons fully validate the effectiveness of our proposed model.

\begin{table}
  \centering
  \caption{Ablation study on the ratio of masked tokens.}
  \label{tab:ablation_mask}
  \resizebox{3.3in}{!}{
  \begin{tabular}{c|c|cc|cc}
    \hline  \toprule
    \multirow{3}{*}{\raggedright \textbf{Masked Ratio}} &\multirow{3}{*}{\raggedright \textbf{Dataset}} &\multicolumn{4}{c}{\textbf{Action Recognition}} \\
    \cline{3-6}
    & &\multicolumn{2}{c|}{\textbf{RGB+Event}} &\multicolumn{2}{c}{\textbf{Event}} \\
    \cline{3-6}
    & & Top1 & Top5 & Top1 & Top5 \\
    \hline 
    0.25 &\multirow{4}{*}{\raggedright \textbf{HARDVS}} &49.35 &59.60   &47.01 &59.03   \\
    0.50 & &51.27 &62.52   &51.28 &60.40   \\
    0.75 & &\textbf{53.24} &\textbf{65.13}   &\textbf{53.18} &\textbf{62.88}  \\
    0.85 & &50.78 &61.73   &50.17 &60.11   \\
    \bottomrule
  \end{tabular} }
\end{table}

\begin{figure*}
\center
\includegraphics[width=\textwidth]{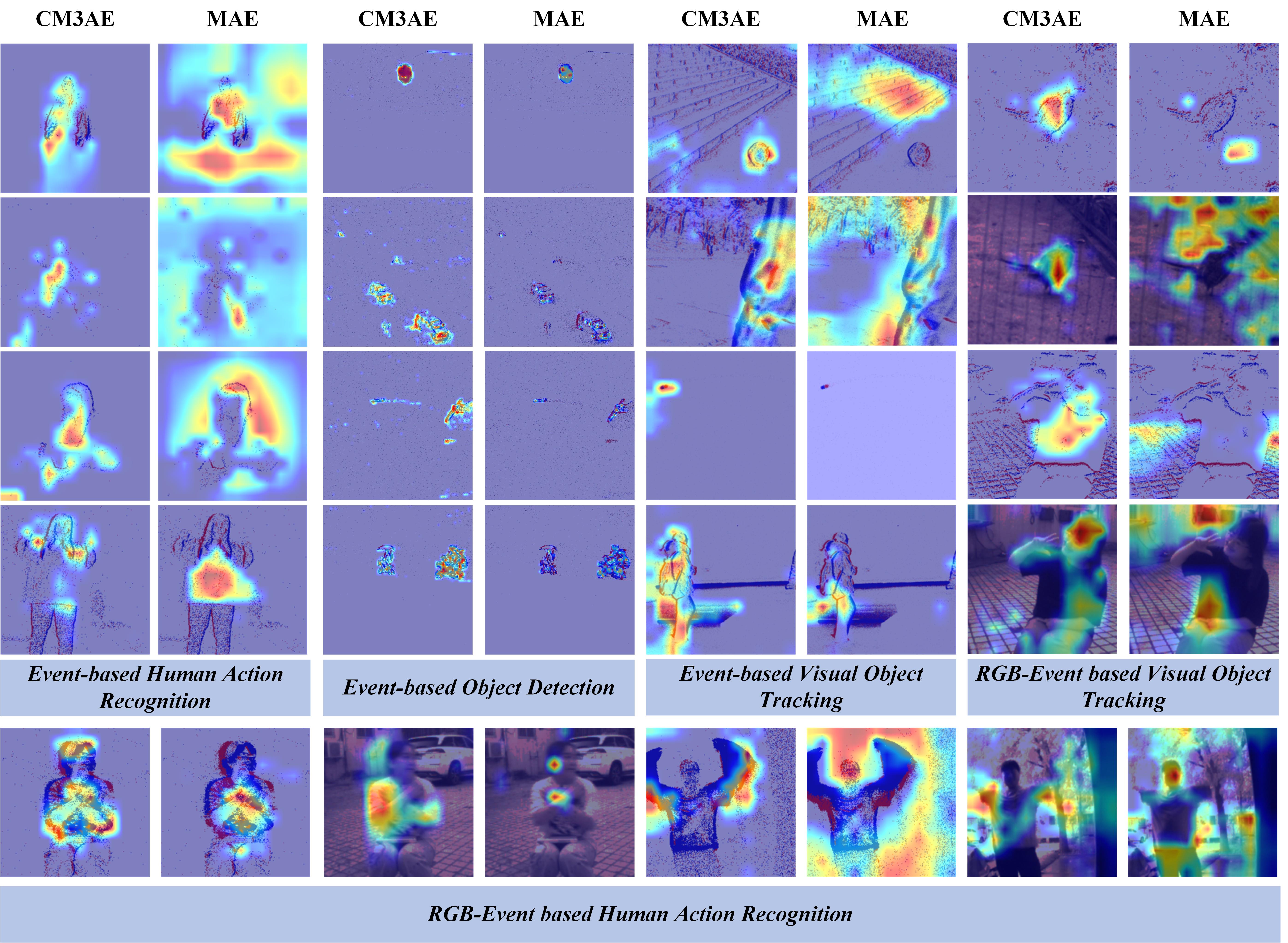}
\caption{Visualization of attention maps on different downstream tasks.}    
\label{fig:attentionmaps}
\end{figure*}

\begin{table*}
  \centering
  \caption{Results of training with only $20\%$ and $10\%$ data in downstream tasks. T.D. denotes training data.}
  \label{tab:ablation_data}
  \resizebox{\textwidth}{!}{
  \begin{tabular}{c|c|c|cc|cc|ccc|ccc|ccc}
    \hline  \toprule
     \multirow{3}{*}{\raggedright \textbf{T.D.}} 
     &\multirow{3}{*}{\raggedright \textbf{Method}} 
     &\multirow{3}{*}{\raggedright \textbf{Dataset}} 
     &\multicolumn{4}{c|}{\textbf{Action Recognition}}
     &\multicolumn{3}{c|}{\textbf{Object Detection}}
     &\multicolumn{6}{c}{\textbf{Visual Object Tracking}}\\
     \cline{4-16} 
     & & &\multicolumn{2}{c|}{\textbf{RGB+Event}} &\multicolumn{2}{c|}{\textbf{Event}} &\multicolumn{3}{c|}{\textbf{Event}} &\multicolumn{3}{c|}{\textbf{RGB+Event}} &\multicolumn{3}{c}{\textbf{Event}}  \\
     \cline{4-16} 
     &  &  & Top1 & Top5 & Top1 & Top5 & $AP_{[0.5:0.95]}$  & $AP_{0.5}$ &  $AP_{0.75}$ & SR & PR & NPR & SR & PR & NPR \\
    \hline 
    \multirow{4}{*}{\raggedright \textbf{$20\%$}} 
     & Scratch & -   &12.94 &28.62   &3.29 &12.15   &34.0 &59.9 &34.4    &29.8 &32.1 &34.3    &28.8 &23.5 &35.8    \\
    & MAE & ImageNet-1K    &20.14 &37.11   &27.32 &48.81   &41.1 &70.1 &42.6    &45.8 &55.2 &56.2    &43.5 &44.9 &57.2   \\
    & MAE & REV2M    &47.37 &58.33   &36.56 &56.12   &41.2 &71.0 &43.2    &47.1 &60.3 &60.5    &46.7 &47.2 &58.9  \\
    & Our & REV2M    &\textbf{49.61} &\textbf{60.73}   &\textbf{44.79} &\textbf{59.31}   &\textbf{43.0} &\textbf{73.2}   &\textbf{44.3}    &\textbf{50.8} &\textbf{63.1} &\textbf{63.0}    &\textbf{48.4} &\textbf{49.8} &\textbf{61.3}     \\

    \hline 
    \multirow{4}{*}{\raggedright \textbf{$10\%$}} 
     & Scratch & -   &5.63 &14.52   &2.40 &8.06   &26.7 &47.3 &26.9    &26.3 &26.9 &30.1    &25.4 &20.0 &30.7    \\
    & MAE & ImageNet-1K    &6.65 &18.04   &9.31 &24.63      &33.3 &60.2 &33.3    &40.1 &48.7 &50.7    &37.0 &35.5 &45.9   \\
    & MAE & REV2M    &33.31 &49.01   &16.94 &37.27   &34.2 &62.1 &34.9    &41.5 &50.8 &52.3    &39.2 &39.8 &48.1   \\
    & Our & REV2M    &\textbf{43.68} &\textbf{57.23}   &\textbf{32.34} &\textbf{52.18}   &\textbf{37.7}   &\textbf{65.9} &\textbf{38.7}    &\textbf{43.9} &\textbf{55.5} &\textbf{56.0}    &\textbf{41.1} &\textbf{42.7} &\textbf{50.5}  \\
    
    \bottomrule
  \end{tabular} }
\end{table*}

\begin{table}
  \centering
  \caption{Ablation study on the fusion module.}
  \label{tab:ablation_fusion}
  \resizebox{3.3in}{!}{
  \begin{tabular}{c|cc|ccc}
    \hline  \toprule
    \multirow{3}{*}{\raggedright \textbf{Fuse Modul}}  &\multicolumn{2}{c|}{\textbf{Action Recognition}} & \multicolumn{3}{c}{\textbf{Visual Object Tracking}}  \\
     \cline{2-6}
     &\multicolumn{5}{c}{\textbf{RGB+Event}} \\
     \cline{2-6}
      & Top1 & Top5 & SR & PR & NPR \\
    \hline 
    w/o Pre-training &52.40 &62.94   &62.0 &74.6 & 73.8  \\
    Pre-training &53.40 &65.82   &62.6 &75.8 &74.9   \\
    
    \bottomrule
  \end{tabular} }
\end{table}

\subsection{Ablation Study} 

\noindent 
\textbf{Effects of Multimodal Fusion Reconstruction Module.}
In this paper, to better model the relationship between RGB and Event modalities, we introduce multimodal fusion reconstruction module(MFRM) that reconstructs RGB images using multimodal fusion features. As shown in Table~\ref{tab:ablation_model}, the introduction of this module during pre-training on the HARDVS dataset leads to performance improvements in both Event-based and RGB-Event based human action recognition tasks. Specifically, when using only the DMA, the Event-based approach achieves Top1 and Top5 accuracies of $49.47\%$ and $60.27\%$, respectively. After further integrating the MFRM into pre-training, the Event-based approach shows improvements of $2.73\%$ in Top1 and $1.78\%$ in Top5 accuracy. The experimental results fully validate the effectiveness of the proposed multimodal fusion reconstruction module.

\noindent 
\textbf{Effects of Multimodal Contrastive Learning.}
The process of reconstructing the original image in an autoregressive manner focuses on modeling local relationships. To enhance the model's ability to capture global dependencies and understand multimodal information, we introduce 
multimodal contrastive learning(MCL) strategy. As shown in Table~\ref{tab:ablation_model}, after introducing MCL based solely on the DMA, the Top1 and Top5 metrics for the Event-based approach increase to $51.49\%$ and $61.16\%$, respectively. When the MFRM and MCL are jointly employed, the Top1 and Top5 metrics for the Event-based approach further increase to $53.18\%$ and $62.88\%$. Similar results are also obtained for the RGB-Event-based approach. These results clearly validate the effectiveness of the proposed multimodal contrastive learning.

\noindent 
\textbf{Analysis on Ratio of Masked Tokens.}
As shown in Table~\ref{tab:ablation_mask}, we conduct pre-training on HARDVS with four different ratios of masked tokens to examine their impact, specifically 0.25, 0.50, 0.75, and 0.85. The results from the Event-based and RGB-Event based human action recognition tasks indicate that the best performance is achieved when $75\%$ of the input tokens are randomly masked.

\noindent 
\textbf{Analysis on Different Sizes of Training Data in Downstream Tasks.}  
In this work, to validate the effectiveness of the CM3AE pre-training model on few-shot learning tasks, we conduct experiments using only $10\%$ and $20\%$ of the training data from downstream tasks. As shown in Table~\ref{tab:ablation_data}, for RGB-Event based human action recognition, when training from scratch with only $10\%$ of the data, the Top1 and Top5 accuracies are $5.63\%$ and $14.52\%$, respectively. After loading a general MAE pre-trained model, the performance improve to $6.65\%$ and $18.04\%$, indicating that generic pre-trained models struggle with this challenging task. When using our integrated pre-training dataset to train the MAE, the performance increases significantly to $33.31\%$ and $49.01\%$. Finally, when loading the CM3AE pre-trained model, the optimal results of $43.68\%$ and $57.23\%$ are achieved. These experimental results and comparisons fully demonstrate the effectiveness of our pre-training model.

\noindent 
\textbf{The impact of the pre-trained fusion module on the performance of downstream tasks.}  
In our pre-training framework, a fusion module is designed to integrate complementary information between the RGB-Event bimodal modalities, helping the model better reconstruct the original image. This module is a simple Transformer Block, which can be easily adapted to existing models for multimodal feature fusion. In Table~\ref{tab:ablation_fusion}, we conduct experiments on RGB-Event-based action recognition and visual object tracking tasks to evaluate the impact of loading pre-trained parameters for the fusion module. In the action recognition task, without loading the pre-trained parameters, the Top1 and Top5 metrics are $52.40\%$ and $62.94\%$, respectively. When the pre-trained parameters are loaded, the results improve to $53.40\%$ and $65.82\%$. Similar results are obtained in the visual object tracking task. These experiments and comparisons validate the effectiveness and generalizability of the fusion module trained through self-supervised learning on large-scale paired multimodal data.

\subsection{Visualization} 

\begin{figure}[!htp]
\center
\includegraphics[width=3in]{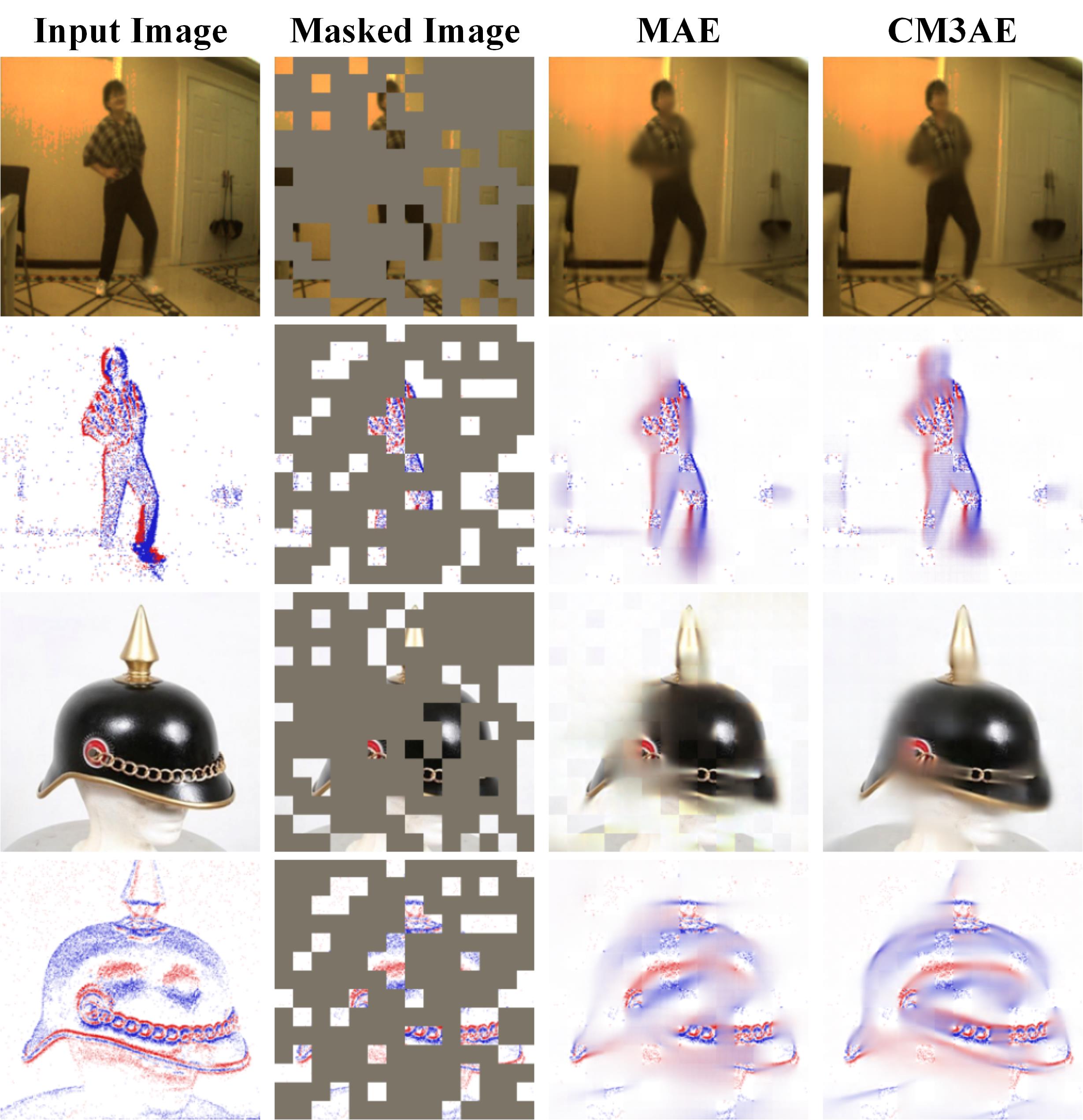}
\caption{Visualization of reconstructed RGB and Event images using MAE and our proposed CM3AE.}  
\label{fig:reconst_vis}
\end{figure}

In this section, we visualize the reconstructed images, along with feature maps and results from different downstream tasks using both MAE and our proposed CM3AE pre-trained model. As shown in Fig.~\ref{fig:attentionmaps}, we apply GradCAM to visualize attention maps from the 11th Transformer layer across five downstream tasks. Compared to MAE, our CM3AE exhibits more distinct heatmap responses in key regions, demonstrating superior performance in both RGB-Event and Event-based tasks. Fig.~\ref{fig:downstream_vis} presents the comparative visualization results between our model and the baseline MAE on five downstream tasks. In Fig.~\ref{fig:reconst_vis}, we present the reconstruction results of MAE and CM3AE on the original images, where the third column displays the MAE reconstructions and the fourth column displays the reconstruction results of our proposed CM3AE. The results demonstrate that our model achieves better reconstruction for both RGB and Event images.

\section{Conclusion}
\label{sec:con}

In this paper, we propose a universal pre-training framework called CM3AE for RGB frames and Event-Voxel/-Frame. In this framework, we first feed RGB and Event images into a dual-stream masked autoencoder, dividing them into non-overlapping patches and randomly masking $75\%$ of them. Notably, during masking, we ensure that half of the unmasked patches in RGB and Event modalities are positionally aligned. The unmasked patches are then processed by modality-specific Transformer encoders and decoders to reconstruct the pixel values of the masked patches. To enhance the model's understanding of multimodal information, we first design a multimodal fusion reconstruction module that reconstructs the original image using fused multimodal features, thereby improving the model's ability to extract complementary information across modalities. Second, we propose a multimodal contrastive learning strategy that leverages the paired relationship between RGB and Event data to strengthen the model's capacity for global representation learning of each modality. Additionally, we construct REV2M, a large-scale pre-training dataset comprising 2.53 million RGB-Event pairs. Extensive experiments on three downstream tasks, human action recognition, object detection, and visual object tracking, fully validate the effectiveness and advantages of the CM3AE framework.
{
    \small
    \bibliographystyle{ieeenat_fullname}
    \bibliography{main}
}


\end{document}